\documentclass[nohyperref]{article}

\usepackage{microtype}
\usepackage{graphicx}
\usepackage{subfigure}
\usepackage{booktabs} 

\usepackage{hyperref}

\usepackage[accepted]{icml2022}

\usepackage{amsmath}
\usepackage{amssymb}
\usepackage{mathtools}
\usepackage{amsthm}
\usepackage{nicefrac}

\usepackage[capitalize,noabbrev]{cleveref}

\theoremstyle{plain}
\newtheorem{theorem}{Theorem}[section]

\theoremstyle{definition}
\newtheorem{definition}[theorem]{Definition}

\theoremstyle{remark}

\newcommand{\Tau}{\mathrm{T}}

\usepackage{balance}

\usepackage[textsize=tiny, prependcaption]{todonotes}

\begin{document}

\twocolumn[
\icmltitle{A Human-Centric Assessment Framework for AI}




\begin{icmlauthorlist}
\icmlauthor{Sascha Saralajew}{nle}
\icmlauthor{Ammar Shaker}{nle}
\icmlauthor{Zhao Xu}{nle}
\icmlauthor{Kiril Gashteovski}{nle}
\icmlauthor{Bhushan Kotnis}{nle}
\icmlauthor{Wiem Ben Rim}{nle}
\icmlauthor{J\"urgen Quittek}{nle}
\icmlauthor{Carolin Lawrence}{nle}
\end{icmlauthorlist}

\icmlaffiliation{nle}{NEC Laboratories Europe GmbH, Heidelberg, Germany}

\icmlcorrespondingauthor{Sascha Saralajew}{sascha.saralajew@neclab.eu}
\icmlcorrespondingauthor{Carolin Lawrence}{carolin.lawrence@neclab.eu}

\icmlkeywords{Human-centric AI, explainable AI, assessment, validation}

\vskip 0.3in
]



\printAffiliationsAndNotice{}  

\begin{abstract}
    With the rise of AI systems in real-world applications comes the need for reliable and trustworthy AI\@. An essential aspect of this are explainable AI systems. However, there is no agreed standard on how explainable AI systems should be assessed. Inspired by the Turing test, we introduce a human-centric assessment framework where a leading domain expert accepts or rejects the solutions of an AI system and another domain expert. By comparing the acceptance rates of provided solutions, we can assess how the AI system performs compared to the domain expert, and whether the AI system's explanations (if provided) are human-understandable. This setup---comparable to the Turing test---can serve as a framework for a wide range of human-centric AI system assessments. We demonstrate this by presenting two instantiations: (1) an assessment that measures the classification accuracy of a system with the option to incorporate label uncertainties; (2) an assessment where the usefulness of provided explanations is determined in a human-centric manner.
\end{abstract}

\section{Introduction}\label{sec:introduction}

AI systems have matured and are on the rise to become an integral part of the real world in applications that span across our entire society. The performance of such AI systems is mostly validated in terms of accuracy against a labeled ground-truth dataset. Even if this is often appropriate, it poses the challenge that such validation frameworks cannot be transferred directly to validate AI systems that provide solutions in terms of a prediction and an explanation, or that exceed human performance. The problem of how to validate explainability methods is vividly discussed and investigated, leading to diverse frameworks---for instance, the concepts of meta-predictor \cite{Fel2021} and simulatability \cite{DoshiVelez2017a} are only proxies that cannot measure an AI system's performance \textit{in comparison to} a human expert.

We describe a generic framework to assess AI systems in a blind experiment, where three domain experts interact in a collaborative environment. One domain expert is a human lead expert, who picks the tasks to be solved and accepts or rejects the provided solutions. Next, each task is solved by a domain expert, either a human or an AI system, whereby the leading expert has no information about who solved the task nor that an AI system might have solved it. Our framework assesses the performance of the AI system compared to the human expert, by estimating the chances that the lead expert accepts a solution provided by either the human or the AI system.

Consider, for example, the assessment of medical laboratories: a leading laboratory (maybe hired by some authority) sends test specimens (the tasks) anonymously to another laboratory. After analyzing the specimens, the laboratory returns the results (solutions). The leading laboratory evaluates the results knowing the sent specimens and reports the acceptance rate of the assessment. What the leading laboratory does not know is that the specimens are analyzed by either a human expert or fully automatized by a machine so that the acceptance rate refers to the human or the system. By comparing the acceptance rate of the human with the system, the quality of the system is assessed. This setup allows an unbiased validation of whether or not it is acceptable to have a machine perform the analysis in place of a human.

In the following, we will describe the proposed assessment framework in detail. Next, to demonstrate the generalizability of this framework, we show how the ordinary measure of classification accuracy emerges from a specific instantiation of the framework and allows us to measure label uncertainty. Additionally, we describe an instantiation to assess the usefulness of AI explainability methods by designing a setup where the lead expert requires an explanation to make a proper assessment in a short amount of time---thus, this setup measures the usability of an explanation method.

The outline of the paper is as follows: The next section defines and discusses the assessment framework and introduces two instantiations as examples. Then, we discuss related work and finish with a conclusion and an outlook.

\section{Assessment Framework}

\begin{figure}
    \centering
    \includegraphics{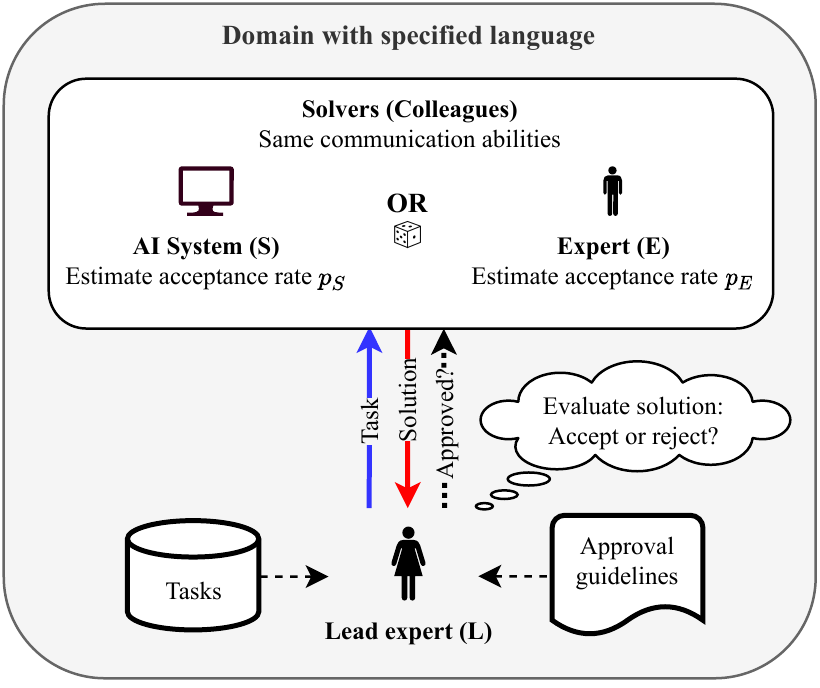}
    \caption{Illustration of the proposed assessment framework. The test is performed in a specific domain with a specified language. The tasks of the leading expert are randomly assigned to either the AI or the expert. Based on the received solution, the lead expert evaluates it. We assess the AI by calculating the acceptance rates.}
    \label{fig:assessment-framework}
\end{figure}

The proposed assessment framework can measure how well an AI system performs a task compared to a human expert. First, we give a formal definition followed by a discussion. Second, we outline two instantiations of the framework.

\subsection{Formal Definition}

Consider the situation in \cref{fig:assessment-framework}. Our assessment framework consists of three \textit{domain experts} (or groups of experts): a \textit{Lead expert} (L), an \textit{Expert} (E), and an \textit{AI System} (S). The lead expert L assigns a task via a well-defined communication channel to one of the colleagues (either E or S). The assignment is made at random and L does not know who will solve the task, nor that there are different solvers involved. After the assigned colleague solves the task, the solution is returned to L via a well-defined communication channel. Then, L decides whether to accept or reject the solution based on specified approval guidelines.\footnote{Acceptance means conformity with the approval guidelines. Thus, a rejection does not imply that the individual parts of a solution (for instance, the prediction and the explanation) are incorrect.} Thus, L assesses (evaluates) the solution for the given task, which does \textit{not} necessarily imply that L has to solve the task again. To compute the acceptance rate, the decision whether the solution is accepted is mapped to the colleague who solved the task (the solver does not know whether their solution was accepted). By repeating the test for several tasks of the domain, we can estimate the acceptance rates for E and S\@.

\begin{definition}
	For a system $S: \Tau \rightarrow \Sigma$, an expert $E: \Tau \rightarrow \Sigma$, and a lead expert $L: \Tau \times \Sigma \rightarrow \left\{0,1\right\}$, the \textit{assessment} consists of determining the empirical probabilities that solutions $\sigma \in \Sigma$ for tasks $\tau \in \Tau$ that are randomly drawn by the lead expert L and are randomly solved by the system S or the expert E are accepted by the lead expert L:
    \begin{align*}
	    p_S &= \frac{1}{\#\Tau_S}\sum_{\tau \in \Tau_S} L\left(\tau, S(\tau)\right),\\
	    p_E &= \frac{1}{\#\Tau_E}\sum_{\tau \in \Tau_E} L\left(\tau, E(\tau)\right),
	\end{align*}

	where the individual task sets $\Tau_S$ and $\Tau_E$ are subsets of the task set $\Tau$ and $p_S$ is the empirical probability that a solution provided by the system S will be accepted by the lead expert~L (analogous interpretation for $p_E$).
\end{definition}

The following outcomes are possible: (1) The AI system performs worse than the expert if $p_S \ll p_E$; (2) The AI system behaves like the expert if $p_S \approx p_E$; (3) The AI system exhibits superhuman abilities if $p_S \gg p_E$.

Note that the assessment of a medical lab mentioned in \cref{sec:introduction} can be mapped to the assessment framework definition. Moreover, the framework is unbiased and human-centric. Unbiased in the sense that the lead expert does not know that there is an AI involved and, thus, evaluates solutions from a human-centric perspective. Additionally, by always involving a human and an AI for task solving, it is required to define how to solve a task and how to communicate with L, which makes the task description and solution communication human-centric as well. For explainable AI, this postulate immediately disqualifies explanation methods that produce explanations that are not suited for human interpretation. Therefore, with a common acceptance of our framework, future explainable AI research can consider how human-centric an explanation method is during its early conception. This is desired as explanations are generated for the sole purpose of being useful for humans. Finally, because the framework always provides a human baseline performance through E, it can quantify superhuman performance.

\subsection{Assumptions, Remarks, and Discussion}


\paragraph{Domain, language, tasks, and solutions:} 
The test is fixed to a certain domain with experts, and the communication is limited to understanding tasks ($\tau \in \Tau$) and solutions ($\sigma \in \Sigma$). These communications require the languages to be well-defined so that all three parties can understand tasks and solutions, and that E and S can formulate solutions in an \textit{unimpeded} manner. Namely, E and S can communicate well with L using the same languages, and L cannot determine which party is providing a solution based on the language used. At the same time, this ensures that a human can understand the explanation produced by S.

Additionally, for each domain, the task must be well-defined so that the criteria for its completion are unambiguous. In other words, it is obvious what has to be done. For example, in object recognition, annotation guidelines clearly specify what an object is, how to annotate it, and, thus, what solutions are expected. Task definition becomes especially important in the context of explainable AI when the solvers have to return an explanation alongside the prediction because it requires defining the expected explanation (e.\,g., what should be highlighted by a saliency map). Moreover, these definitions set the rules for how E should solve a task to control human subjectiveness. Finally, note the solution language might contain a word for ``no solution derived'' to ensure a solution is always returned even if the AI system encounters errors or the expert cannot provide a solution.

\begin{figure}
    \centering
    \includegraphics[width=0.45\columnwidth]{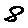}\quad\quad\includegraphics[width=0.45\columnwidth]{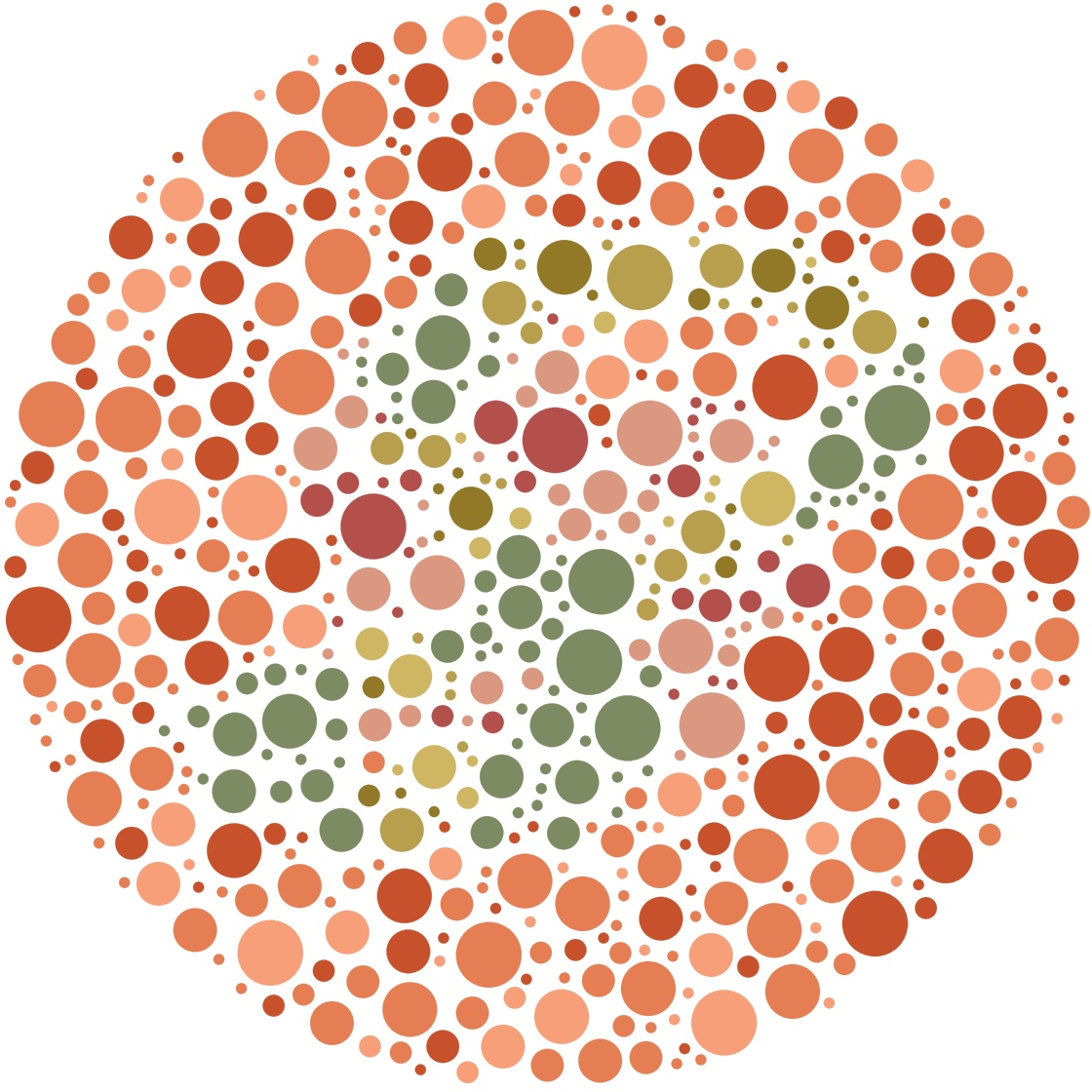}
    \caption{Example of a generated colorblind image (right) of a MNIST image (left) according to plate 4 of Ishihara~\yrcite{Ishihara1972}. People with red-green color deficit would have trouble reading it.}
    \label{fig:colorblind_example}
\end{figure}

\paragraph{Lead expert:} 
The test requires that the lead expert is interested in assessing the solvers by evaluating the solutions following the approval guidelines. If this is not the case, the lead expert could accept any solution, which would lead to the logical consequence that S and E perform equally well because no domain-specific task-solving abilities are required to provide acceptable solutions. 

Importantly, it is not required that L can solve tasks (contrarily to E and S). However, L must be able to evaluate task-solution pairs even if it is time-consuming,  otherwise, the assessment (or validation) of any system is impossible. Consider AlphaFold~\cite{Jumper2021}: protein structures predicted by the model must be evaluated by experiments to confirm correctness. Though time-consuming (but possible), it was used to validate the outstanding model performance.

The approval guidelines are of utmost importance for the evaluation of solutions. Similar to the precise task description (which is related to annotation guidelines), the approval guidelines must specify as precisely as possible how a solution must be evaluated. Every undefined aspect will be impacted by the subjectiveness of the domain lead expert, which can lead to intended or unintended biased evaluations.\footnote{Tasks with known solutions can be injected in the assessment framework to control the compliance with the approval rules of L and task solving rules of E\@.}

\subsection{Assessment of Classification Accuracy}

This example instantiation shows the generalizability of the framework: it can measure the classification accuracy (with label uncertainty) of an AI system S on a given test set $(x_i, y_i) \in D$, where $x_i$ is an input annotated with the class label $y_i$. In the context of the framework, the inputs $x_i$ represent the task set $\Tau$, and the possible class labels $y_i$ form the solution set $\Sigma$ so that the framework assesses the provided class labels of inputs. Additionally, since each $x_i$ was annotated by a human expert, it is feasible to assume that the corresponding label $y_i$ represents the solution of the expert E\@: $E(x_i) = y_i$.
Now, we can define the classification accuracy of a system S \textit{with respect to the lead expert} L by 
\[
    \mathrm{acc}_L(S) = \frac{p_S}{p_E}.
\]

If we further assume that the lead expert L accepts the solution $\sigma_i$ for a task $\tau_i = x_i$ if and only if $\sigma_i = y_i$, then the probability $p_E$ to accept solutions provided by the lead expert E becomes $1.0$, and the classification accuracy with respect to the lead expert $\mathrm{acc}_L(S)$ becomes the canonical classification accuracy $\mathrm{acc}(S)$ used to assess the performance of a system S:
\[
\begin{aligned}
        \mathrm{acc}_L(S) &= \frac{p_S}{p_E} = p_S = \frac{1}{\#\Tau_S}\sum_{\tau \in \Tau_S} L\left(\tau, S(\tau)\right)\\
        &= \frac{1}{\#D}\sum_{i=1}^{\#D} [S(x_i) = y_i] = \mathrm{acc}(S).    
\end{aligned}
\]
If the acceptance criteria of the lead expert L would \textit{not} be the class label $y_i$ of the test input $x_i$ but really an acceptance evaluation of a human expert, then we would naturally identify labels where human experts disagree so that the label uncertainty can be assessed.

\subsection{Assessment of the Usefulness of Image Classification Explanations}

Several researchers investigated the usefulness of explanations in different experimental settings (see \cref{sec:related_work}). To validate whether explanations are useful and help users to assess the correctness of a prediction, we propose an experiment based on the assessment framework with lead experts that have a slight color vision deficit such that they need explanations to assess the predictions for colorblind images derived from MNIST~\cite{LeCun1998}, see \cref{fig:colorblind_example}, \textit{in a short amount of time.} Here, the controlled independent variable is whether an explanation is presented. The dependent variable is the acceptance rate for a given amount of approval time. We determine the usefulness of human-understandable explanations by computing the changes in the acceptance rate between the assessment with and without an explanation. 
This experiment is a suitable instantiation of the framework as it only requires that experts know the Arabic numerals and aptly uses the color perception abilities of humans to assess the usefulness of explainability methods with a reduced experimental bias.

In this instantiation, the AI system S is a neural network with an explainer (e.\,g., an occlusion map by \citealp{Zeiler2013}) that classifies the MNIST colorblind images. Similar to S, the expert E has to provide a prediction and an explanation that highlights where in the image the numeral can be found. To fulfill this task, E must have normal color vision. In contrast, the lead expert L must have a \textit{slight color deficit} such that it is \textit{difficult} for L to see the numeral in a short amount of time---Ishihara~\yrcite{Ishihara1972} specified that humans with normal color vision must see the numeral within 3\,s, whereas humans with a slight color deficit need long exposures to see the numeral. The approval criterion is that L must only accept solutions if L can see the predicted number in the input, which is possible for L to evaluate because L is chosen to have only a slight color deficit.

In the first run of the experiment, solutions without explanations are presented. Because L has a color deficit, the acceptance rates for a short decision time will be low for both E and S\@.\footnote{Given a decision time, each accepted solution where the decision took longer will be counted as rejection internally.} In the second run, each solution includes an explanation. If the explanation is human-understandable, it will help L see the numeral so that the acceptance rates for a short decision time will increase. Therefore, by computing the differences between the run with and without explanation for a short decision time the usefulness of an explanation can be assessed because without explanations, L needs a longer time to evaluate task-solution pairs (L cannot circumvent the need for explanations to achieve short decision times since L needs long exposures to solve the tasks). Moreover, by comparing the acceptance rates, the explanation quality of S compared to E can be assessed, and, by repeating the experiment with different explanation methods, the quality of explanation methods can be quantified.

\section{Related Work}\label{sec:related_work}

The assessment framework we propose builds on the idea of the Feigenbaum test~\cite{Feigenbaum2003}, which is a refinement of the Turing test~\cite{Turing1950}, where the test is set up as a game that is played between experts of a particular (narrow) domain. In this game, a judging domain expert poses, for instance, problems, questions, or theories, which are passed on via two channels to either a computer or another domain expert. The judging domain expert does not know which channel connects to the computer. Depending on the channel, either the computer or the other domain expert replies with an answer. The test asks the following question: by evaluating the received answer, can the domain expert determine which channel connects to the computer? Similar to the Turing test, the Feigenbaum test is a behavioral test that tries to ``test the facet of quality of reasoning''~\citep[p.\,36]{Feigenbaum2003}. For a computer program to pass the test, it must be able to simulate human \textit{intelligent} behavior, which is why the test is sometimes inappropriately taken as a test of human intelligence. We follow the idea of performing an experiment between experts of a certain domain but modify it by proposing a framework where the chances of accepting a solution (answer) from the machine and the human expert are measured. Consequently, the proposed framework is not a test that can be passed, but rather an assessment of solutions for domain-specific tasks so that a computer's performance can be quantified in comparison with human performance. 

To quantify whether or not explanations are human-like, Biessmann and Treu~\yrcite{Biessmann2021} created a Turing test for transparency to evaluate whether humans can identify who generated an explanation (an AI or a human). Since they draw inspiration from the Turing test, this concept is similar to our framework. However, our goal is to assesses any performance of an AI system and a human expert---not only how human-like explanations are. Furthermore, their framework requires the interrogator to be informed about the presence of an AI and a human so that the interrogator may be biased against the AI~\cite{Dietvorst2015}. Our proposal avoids this potential bias.

Other concepts to evaluate explanations is simulatability \citep[given the input and the corresponding explanation, the model output has to be predicted]{DoshiVelez2017a}, and the Meta-predictor~\citep[after a training phase, humans have to predict the model output only by seeing the input]{Fel2021}. 
Hase and Bansal~\yrcite{Hase2020a} performed controlled experiments to measure simulatability, which is conceptually similar to the work of Fel et~al.~\yrcite{Fel2021}. Based on the results, in both experiments, the authors concluded that some explainability methods help users. Similar to our proposed framework, both require two trials (with and without explanation) to measure the usefulness of an explainability method. 
But, with both concepts it is not possible to analyze whether a model is judged to be bad due to superhuman model capabilities, since the concepts are limited to the mental abilities of the human subjects who have to simulate the model behavior.

Alufaisan et~al.~\yrcite{Alufaisan2021} also performed an experiment to evaluate the impact of explanations to help users perform a prediction and concluded that explanations do not positively impact prediction accuracy of humans. However, this result could be affected by uncontrolled confounders like asking the users for a prediction and giving them the freedom to ignore the AI outputs, which is resolved in our framework.

\section{Conclusion and Outlook}

The growing field of explainable AI still has no unified evaluation framework for explainability methods. Based on the contributions of several other frameworks and their experiments, we proposed an assessment framework that combines several of these approaches and addresses their weaknesses. Notably, the proposed framework is human-centric and able to identify models with superhuman performances because it always compares the AI performance with a human baseline performance. To demonstrate the generalizability of the framework, we have described two instantiations: the first measures classification accuracy, and the second measures the usefulness of human understandable explanations. The next steps will be the implementation of the second experiment.

\bibliography{references_db}
\bibliographystyle{icml2022}

\end{document}